\documentclass[conference]{IEEEtran}
\IEEEoverridecommandlockouts
\usepackage{cite}
\usepackage{amsmath,amssymb,amsfonts}
\usepackage{algorithmic}
\usepackage{graphicx}
\usepackage{textcomp}
\usepackage{xcolor}
\usepackage[linesnumbered,ruled,vlined]{algorithm2e}
\usepackage{booktabs} 
\usepackage{newfloat}
\usepackage{listings}
\usepackage{bibentry}
\def\BibTeX{{\rm B\kern-.05em{\sc i\kern-.025em b}\kern-.08em
    T\kern-.1667em\lower.7ex\hbox{E}\kern-.125emX}}
\begin{document}

\title{Label-guided Facial Retouching Reversion\\
\thanks{\IEEEauthorrefmark{1}Equal contribution. \IEEEauthorrefmark{2}Corresponding author (zhangjian.sz@pku.edu.cn). This work was supported in part by National Natural Science Foundation of China (No. 62372016), Guangdong Provincial Key Laboratory of Ultra High Definition Immersive Media Technology (No. 2024B1212010006).}
}

\author{\IEEEauthorblockN{Guanhua Zhao$^1$\IEEEauthorrefmark{1},
Yu Gu$^{1}$\IEEEauthorrefmark{1},
Xuhan Sheng$^{1}$,
Yujie Hu$^{1}$,
Jian Zhang$^{1,2}$\IEEEauthorrefmark{2}
}
\IEEEauthorblockA{$^1$School of Electronic and Computer Engineering, Peking University, China}
\IEEEauthorblockA{$^2$Guangdong Provincial Key Laboratory of Ultra High Definition Immersive Media Technology,}
\IEEEauthorblockA{Shenzhen Graduate School, Peking University, China}
}


\maketitle

\begin{abstract}

With the popularity of social media platforms and retouching tools, more people are beautifying their facial photos, posing challenges for fields requiring photo authenticity.
To address this issue, some work has proposed makeup removal methods, but they cannot revert images involving geometric deformations caused by retouching. To tackle the problem of facial retouching reversion, we propose a framework, dubbed Re-Face, which consists of three components: a facial retouching detector, an image reversion model named FaceR, and a color correction module called Hierarchical Adaptive Instance Normalization (H-AdaIN). 
FaceR can utilize labels generated by the facial retouching detector as guidance to revert the retouched facial images. Then, color correction is performed using H-AdaIN to address the issue of color shift. Extensive experiments demonstrate the effectiveness of our framework and each module. 
\end{abstract}


\section{Introduction}
\label{sec:intro}

In contemporary society, we witness the increasing prevalence of photo editing tools and techniques, enabling virtually anyone to digitally construct a version of themselves closer to their ideals with a simple click or swipe~\cite{rathgeb2022handbook}. 
While these technologies can positively influence personal branding and beauty pursuits, they present challenges in contexts requiring authenticity. Facial recognition systems rely on accurate images in fields such as judicial crime investigation, financial services, and security supervision. Retouched photos can compromise these systems' accuracy, leading to erroneous identity confirmation and security risks~\cite{Dantcheva_Chen_Ross_2012}. Similarly, in online social platforms involving interpersonal trust, excessively retouched photos may violate the principles of honest communication, leading to social issues such as identity fraud. Therefore, developing facial retouching reversion techniques becomes crucial to ensure the integrity and reliability of these systems.

Facial retouching reversion aims to revert images that have undergone retouching processes using software such as Photoshop or Meitu to their original states. The range and intensity of retouching methods applied by different software and users vary widely, making it difficult to precisely detect and control the range and intensity of reversion involving retouching methods, posing significant challenges for the reversion task. To tackle these challenges, we propose Label-driven Facial Retouching Reversion (\textbf{Re-Face}).

Firstly, we employ a facial retouching detector~\cite{Ying_Liu_Li_Xu_Qian_Zhang_2023} capable of identifying the retouching methods and degrees presented in retouched images. We select a modest ensemble of retouching techniques, comprising eye enlarging, face lifting, and smoothing. The facial retouching detector predicts a retouching label containing three integers, indicating the retouching methods and degrees. Then, the face reversion network (\textbf{FaceR}) reverts the retouched image guided by the retouching label. The Latent Diffusion Models (LDM)~\cite{Rombach_Blattmann_Lorenz_Esser_Ommer_2022} trained on the large-scale image dataset ImageNet~\cite{Russakovsky_Deng_Su_Krause_Satheesh_Ma_Huang_Karpathy_Khosla_Bernstein_etal._2015} possess rich pre-trained priors and high-quality generation characteristics. Utilizing LDM, we can ensure the fidelity and consistency of the reverted images. To manipulate the image reversion process using the retouching label, we may directly employ the label as a condition to control the denoising process of LDM. However, the direct fine-tuning of LDM may compromise its robust image priors, thereby impairing the reversion outcomes. Consequently, we train an additional module FaceNet to learn the correspondence between the retouching label and the low-level information of the image, while keeping the LDM frozen.

Additionally, we observe color distribution shifts in the generated reverted images, which is also mentioned in \cite{Choi_Lee_Shin_Kim_Kim_Yoon}. Adaptive Instance Normalization (AdaIN)~\cite{Huang_2017_ICCV} 
is used to control color and stylistic consistency in images. 
However, we find that AdaIN to the entire image is not effective in addressing the issue of color shift. Therefore, we propose Hierarchical Adaptive Instance Normalization (\textbf{H-AdaIN}), which extends global adaptive normalization to hierarchical adaptive normalization computations, enabling faithful color reversion of generated images to match the original images' colors. 

Our contributions can be summarized as follows:

1. We propose a framework, \textbf{Re-Face}, that employs retouching labels generated by the facial retouching detector as conditions to achieve facial retouching reversion. Furthermore, to the best of our knowledge, our Re-Face is the first endeavor to realize facial reversion from deformation retouching. 

2. We introduce Hierarchical Adaptive Instance Normalization (\textbf{H-AdaIN}), effectively addressing color shift issues arising from diffusion models.

3. Extensive analysis and experiments demonstrate the effectiveness of our framework and each module.

\section{Related work}

\subsection{Makeup Removal}

Makeup removal is the task of reverting the original facial appearance from facial images with makeup while maintaining consistency. Previous works learning to achieve makeup transfer and removal~\cite{Chen_Shen_Jia_2017,li2018beautygan,Gu_Wang_Chiu_Tai_Tang_2019,Liu_Jiang_Gao_He_Feng_Li_Yan_2021,sun2022ssat}. 
These works can only handle non-deformative makeup removal tasks and do not apply to retouching reversion tasks involving deformations.

\begin{figure*}[t]
\centering
\includegraphics[width=\textwidth]{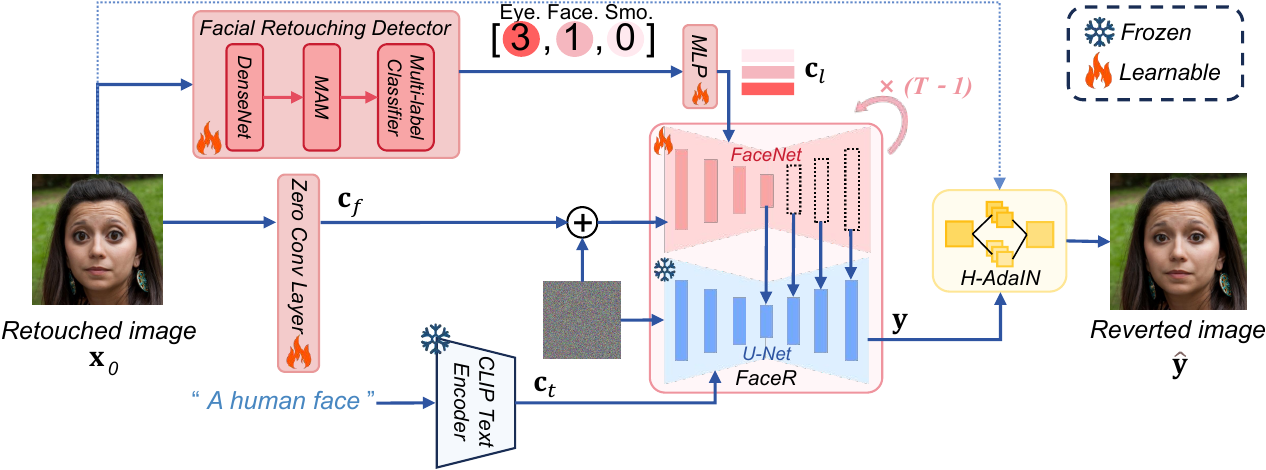}
\caption{Overall framework of the proposed Re-Face. The overall architecture is composed of three parts: 1) a facial retouching detector; 2) an image reversion model -- FaceR, which is composed of a pre-trained LDM with fixed parameters and a trained FaceNet; 3) a color correction module -- H-AdaIN.}
\label{fig: overall}
\end{figure*}

\subsection{Facial Retouching Detection}
\label{related_detection}
~\cite{bharati2016detecting,Sharma_Singh_Goyal_2023,Jain_Majumdar_Singh_Vatsa_2020,Rathgeb_Satnoianu_Haryanto_Bernardo_Busch_2020} classify facial images as either original or retouched. These works can only detect whether an image has been retouched or not, but cannot detect the specific methods and degrees of retouching. Additionally, there are specialized detection networks designed to detect retouched images~\cite{Jain_Singh_Vatsa_2018}, as well as detect occluded faces~\cite{Geng_Peng_Huang_Tian_2020}.

\subsection{Diffusion-based Generative Models}
The diffusion models~\cite{Sohl-Dickstein_Weiss_Maheswaranathan_Ganguli_2015, Song_Ermon_2019, Ho_Jain_Abbeel_2020} gain success in visual generation. Although text-to-image (T2I) diffusion models possess powerful capabilities for generating high-quality images from text prompts, natural language inherently lacks the fine-grained control needed for detailed image synthesis. To address this limitation, various methods have been proposed to integrate conditional control into T2I diffusion models~\cite{ho2022classifier, zhang2023adding, Kumari_Zhang_Zhang_Shechtman_Zhu_2022, zhang2023controllable, qin2023unicontrol, mou2024dragon, mou2024diffeditor}. In this paper, we customize an image-conditioned diffusion model and demonstrate that it significantly enhances adaptability to reverse the retouched facial images.

\section{Method}

In this section, we introduce a novel framework -- Re-Face shown in Fig. \ref{fig: overall}.

\subsection{Facial Retouching Detector}
\label{sec:FRD}
In the task of retouched image reversion, it is necessary to detect the retouching methods and their corresponding degrees to accurately revert the retouched image. The reversion network, FaceR, can follow the guidance of the detection results for reversion, rather than blindly reverting. 
Considering that our task requires fine-grained detection of retouching methods and degrees, we choose DenseNet121-MAM~\cite{Ying_Liu_Li_Xu_Qian_Zhang_2023}, as shown in Fig.~\ref{fig: overall}. It is an adaptive token clustering method based on DenseNet121~\cite{huang2017densely}. It utilizes attention mechanisms and Gumble Softmax~\cite{jang2017categorical} to perform token clustering and reduce spatial redundancy through mean projection. Finally, a multi-granularity attention mechanism is adopted to connect reduced token structures and predict different levels of retouching operations through a multi-label classifier. The entire detection process can be defined as follows:
\begin{equation}
    \mathbf{m}_\textit{l} = FRD(\mathbf{x}_\textit{0}),
\end{equation}
where $\mathbf{m}_\textit{l}$ represents a retouching label with 
dimensions $1 \times 3$, while $\mathbf{x}_\textit{0}$ denotes the input retouched image. For the input $\mathbf{x}_\textit{0}$.

\subsection{FaceR}
\label{sec:FaceR}
FaceR is a key component of the framework Re-Face, aiming to accurately revert a retouched image based on the retouching label $\mathbf{m}_\textit{l}$. 
We employ a \textbf{FaceNet} to learn the reversion of retouched images, while also leveraging the rich prior knowledge provided by LDM to generate high-quality images. At the same time, this approach avoids overfitting and catastrophic forgetting. LDM trains an autoencoder $(\mathcal{E})$ capable of transforming pixel-space images into latent code $\mathbf{z}_\textit{0} = \mathcal{E}(\mathbf{x}_\textit{0})$ and decoding them back into pixel space through the decoder. In addition, LDM trains an improved U-Net denoiser to directly denoise in the latent space. The optimization process can be defined by the following equation:
\begin{equation}
    \mathcal{L}=\mathbb{E}_{\mathbf{z}_\textit{t},\mathbf{C},t,\mathbf{\epsilon}\sim\mathcal{N}(0,I)}(||\mathbf{\epsilon}-\mathbf{\epsilon}_\theta(\mathbf{z}_\textit{t},t,\mathbf{C})||_2^2),
\end{equation}
where $\mathbf{z}_\textit{t}$ represents the noise sample of $\mathbf{z}_\textit{0}$ after $t$ time steps, $\mathbf{\epsilon}$ denotes the noise feature map at time step $t$, $C$ denotes the control information, and $\mathbf{\epsilon}_{\theta}$ represents the U-Net denoiser, used to predict the noise added to $\mathbf{z}_\textit{t}$. During inference, deterministic DDIM sampling is employed to denoise the stochastic noise $\mathbf{z}_\textit{t}$ into the latent code $\mathbf{z}_\textit{0}$, which is then decoded into the pixel domain $\mathbf{\hat{x}}_\textit{0}$ through the VAE Decoder. At each time step $t$, to balance sample quality and conditional alignment, a classifier-free guidance method~\cite{Ho_Salimans} is used.
\begin{equation}
    \mathbf{\tilde{\epsilon}}_\theta(\mathbf{z}_\textit{t},t,\mathbf{C})=(1+\omega)\mathbf{\epsilon}_\theta(\mathbf{z}_\textit{t},t,\mathbf{C})-\omega\mathbf{\epsilon}_\theta(\mathbf{z}_\textit{t},t),
\end{equation}
where $\mathbf{\tilde{\epsilon}}_\theta$ represents the score guided by the classifier, used to update $\mathbf{z}_\textit{t}$ towards $\mathbf{z}_\textit{t-1}$. $\omega$ is a scalar controlling the strength of the guidance by $\mathbf{C}$. The FaceNet is a trainable copy of the encoder block and middle block of U-Net in LDM. Its output is added to the 12 skip connections and 1 middle block of U-Net in LDM. FaceR achieves reversion using the conditions $\mathbf{c}_\textit{t}$, $\mathbf{c}_\textit{l}$, $\mathbf{c}_\textit{f}$. Here, $\mathbf{c}_\textit{t}$ is a semantic prompt encoded using the CLIP text encoder~\cite{Radford_Kim_Hallacy_Ramesh_Goh_Agarwal_Sastry_Amanda_Mishkin_Clark_etal._2021} (e.g., ``a human face''), which is input into LDM. We pass the retouching label $\mathbf{m}_\textit{l}$ (e.g., [1,0,3] representing the degrees of eye enlarging, face lifting, and smoothing, respectively) through fully connected layers to obtain an embedding $\mathbf{c}_\textit{l}$, with a shape of 3$\times$768, the process of the MLP is as follows:
\begin{equation}
    \mathbf{c}_\textit{l} = \text{MLP}(\mathbf{m}_\textit{l}) = \text{reshape}(\mathbf{m}_\textit{l} \cdot \mathbf{W}, [-1, 3, 768]),
\end{equation}

\begin{figure}
\centering
\includegraphics[width=0.45\textwidth]{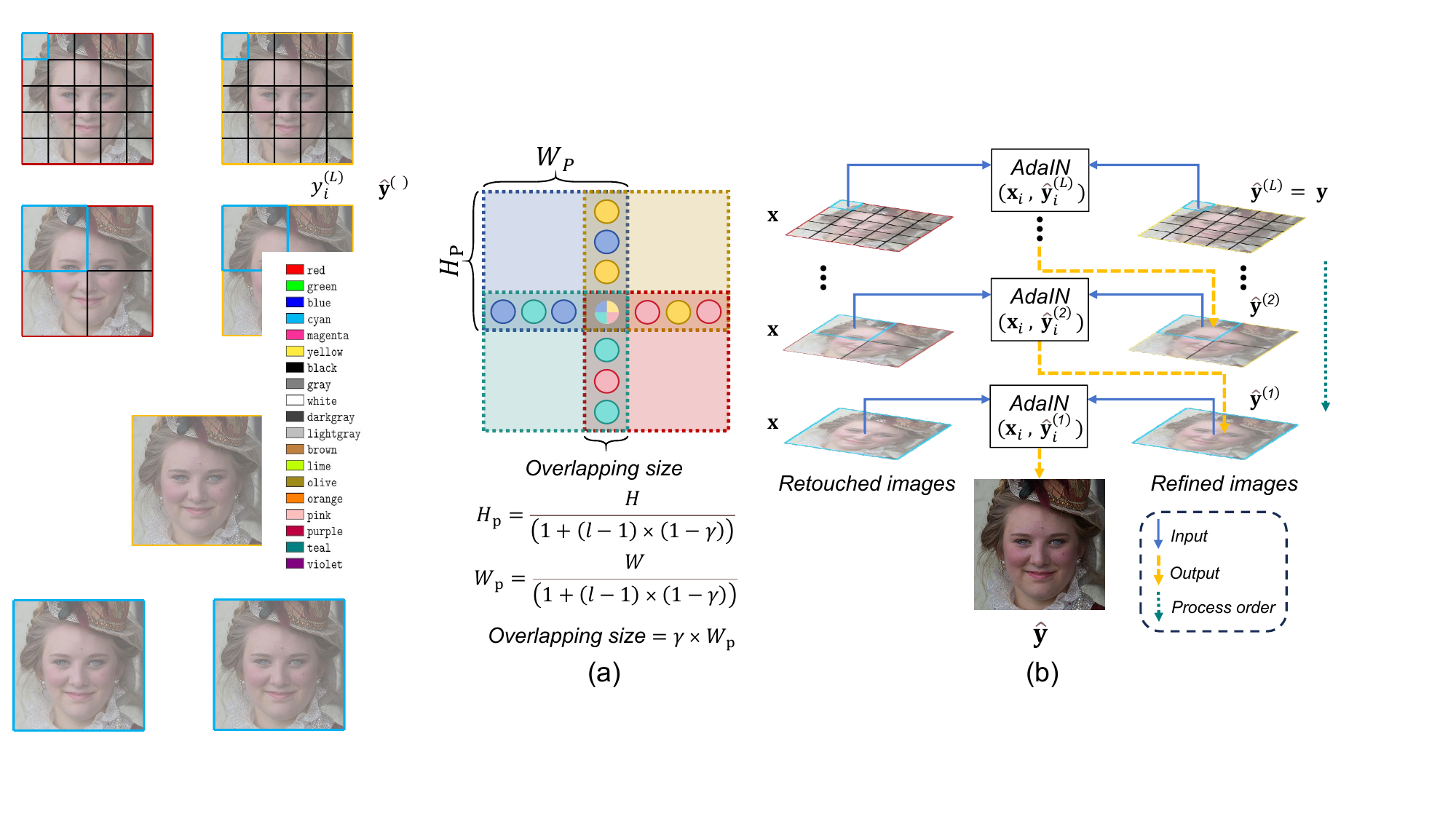}
\caption{Hierachical Adaptive Instance Normalization (H-AdaIN). (a) The formula for overlapping patches. (b) Overview of H-AdaIN. At level $l$, each pair of corresponding $i^{th}$ patches (${\mathbf{x}}_{\textnormal{\textit{i}}}$, $\mathbf{y}_\textnormal{\textit{i}}^{\textnormal{\textit{l}}}$) from $\mathbf{x}$ and $\mathbf{y}^{\textnormal{\textit{l}}}$ are input into AdaIN to obtain current refined result $\mathbf{y}^{\textnormal{\textit{l-1}}}$, 
which is used as input at the next level $l-1$. The process order of AdaINs is performed from the finest level $L$ to the coarsest level $1$. 
} 
\label{fig:hadain}
\end{figure}

\noindent where $\mathbf{W}$ is the weight matrix of the fully connected layer. $\mathbf{c}_\textit{f}$ represents the retouched image reshaped into the latent space, obtained by zero convolution layers. The process of obtaining $\mathbf{c}_\textit{f}$ is as follows:
\begin{equation}
    \mathbf{c}_\textit{f} = \mathcal{Z}(\mathbf{x}_\textit{0};\Theta).
\end{equation}
FaceR implicitly extracts the image information from $\mathbf{c}_\textit{f}$ and edits it according to the $\mathbf{c}_\textit{l}$ to achieve retouching reversion.

\begin{figure*}[t]
\centering
\includegraphics[width=\textwidth]{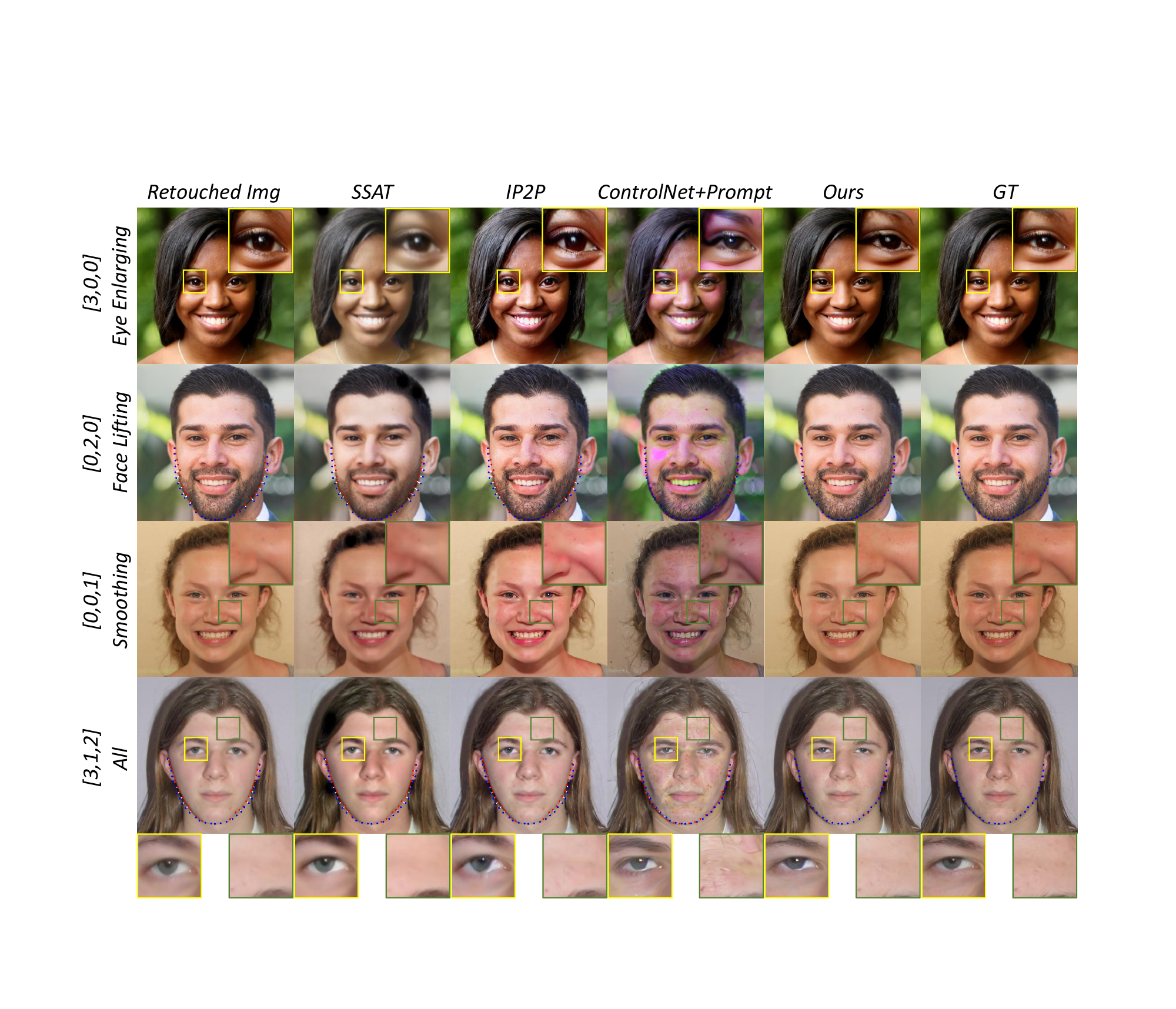}
\caption{ Visual comparison with other methods. The labels to the left of the images indicate the types and degrees of retouching. The blue landmarks represent the GT cheek, while the red ones represent the cheek of the current face. The white solid lines indicate the distance between the corresponding red and blue landmarks. \textbf{Please zoom in for details}.
} 
\label{fig:compare_fig}
\end{figure*}

\subsection{Hierarchical Adaptive Instance Normalization}

\label{sec:H-AdaIN}
\begin{algorithm}
    \SetAlgoLined 
	\caption{Hierarchical Adaptive Instance Normalization} 
        \label{algo:hadain}
	\KwIn{
        \begin{itemize}
            \item hierarchical level $L$ and overlapping ratio $\gamma$,
            \item retouched image $\mathbf{x}\in \mathbb{R}^{C \times H \times W}$,
            \item reverted image $\mathbf{y}\in \mathbb{R}^{C \times H \times W}$.
        \end{itemize}
        } 
	\KwOut{corrected result $\hat{\mathbf{y}}$}
        $\hat{\mathbf{y}}^{\textnormal{\textit{(L)}}}=\mathbf{y}$\\
	\For{$l=L$ to $1$}{
        $ H_{\text{p}} = \lceil \frac{H}{(1+(l-1) \times (1-\gamma))} \rceil $\\
        $ W_{\text{p}} = \lceil \frac{W}{(1+(l-1) \times (1-\gamma))} \rceil $\\
		$\mathbf{x}_\text{p}^{\textnormal{\textit{(l)}}} = \text{Patchify}(\mathbf{x}, H_{\text{p}}, W_{\text{p}}, \gamma) \in \mathbb{R}^{{N}_{\text{p}} \times C \times H_{\text{p}} \times W_{\text{p}}} $ \\
		$\hat{\mathbf{y}}_\text{p}^{\textnormal{\textit{(l)}}} = \text{Patchify}(\hat{\mathbf{y}}^{\textnormal{(\textit{l})}}, H_{\text{p}}, W_{\text{p}}, \gamma)\in \mathbb{R}^{{N}_{\text{p}} \times C \times H_{\text{p}} \times W_{\text{p}}}$ \\
            \For{$i=0$ to $N_\text{p}$}{
                $\tilde{\mathbf{y}}_\text{p}^{\textnormal{(\textit{l}})}[i]=\text{AdaIN}(\hat{\mathbf{y}}_\text{p}^{\textit{(l)}}[i],\ \mathbf{x}_\text{p}^{\textit{(l)}}[i]) \in \mathbb{R}^{C \times H_{\text{p}} \times W_{\text{p}}} $
            }
        $\hat{\mathbf{y}}^{(\textnormal{\textit{l-1}})} = \text{DePatchify}(\tilde{\mathbf{y}}_\text{p}^{\textnormal{\textit{(l)}}}, H, W, \gamma)\in \mathbb{R}^{C \times H \times W}$
	}
        $\hat{\mathbf{y}}=\hat{\mathbf{y}}^{\textnormal{\textit{(0)}}}$ \\
        \Return $\hat{\mathbf{y}}$
\end{algorithm}
Diffusion models often exhibit strong generative capabilities but are frequently plagued by color shifts \cite{Choi_Lee_Shin_Kim_Kim_Yoon}. To address the color mismatch issue between the reverted image and the input conditional image, a straightforward approach is to employ adaptive instance normalization (AdaIN) \cite{stablesr_wang2023exploiting}. Specifically, let $\mathbf{x}\in \mathbb{R}^{C \times H \times W}$ denote the input image to FaceR and $\mathbf{y}\in \mathbb{R}^{C \times H \times W}$ represent the reverted image from FaceR. Then the calculation formula for AdaIN can be expressed as:

\begin{equation}
    \hat{\mathbf{y}}^{c} = \text{AdaIN}(\mathbf{x},\mathbf{y}) = \mu_{\mathbf{x}}^c +  \sigma_{\mathbf{x}}^c \frac{ ( {\mathbf{y}}^{c} - \mu_{\mathbf{y}}^c )}{\sigma_{\mathbf{y}}^c},
\end{equation}
where $\mu$ and $\sigma$ represent estimated mean and standard deviation, respectively, and $c \in \{R, G, B\}$ denotes the three color channels. AdaIN performs global consistent scaling and shifting of all pixel values by computing the statistical features of the input and output images. Although this simple global color correction technique can mitigate color discrepancies between the generated images and input images to a certain extent, when encountering complex backgrounds or accessories around the face, the generated images by the reversion network exhibit not only simple global color shifts but also varying degrees of color shifts in different image regions. As illustrated in Fig. ~\ref{fig:hadaincom}, row 1, column 3, under these circumstances, AdaIN is unable to revert the color of the earring faithfully.

To address the aforementioned new issue, we further propose Hierarchical Adaptive Instance Normalization (H-AdaIN), as illustrated in Fig. \ref{fig:hadain}. We perform different levels of overlapping patchification on the retouched image $\mathbf{x}$ and the initial generated image $\mathbf{y}$ simultaneously. At each level $l$, from $\mathbf{x}$ and $\mathbf{y}^{\textit{(l)}}$, we find the corresponding pair of patches ($\mathbf{x}_\textit{i}$, $\mathbf{y}_\textit{i}^{\textit{(l)}}$), which are fed into AdaIN. The overall process order is from the finest level $L$ to the coarsest level $1$. The details of H-AdaIN can be referred to Algo.~\ref{algo:hadain}.

\begin{table*}
\centering
\setlength{\tabcolsep}{10pt}
\caption{Quantitative results.
"No" indicates the absence of retouching labels, "GT" denotes the usage of real retouching labels, and "Detector" indicates the utilization of retouching labels generated through the facial retouching detector. Red and bold signify the optimal values, and blue denotes sub-optimal values. The AvgD between retouched images and GT images is \textbf{7.13}.}   
\label{tab:comp}
\begin{tabular}{@{}cccccccc@{}}
\toprule
Method            & Label    & FID$\downarrow$    & SSIM$\uparrow$   & PSNR$\uparrow$  & LPIPS$\downarrow$  & DISTS$\downarrow$  & AvgD$\downarrow$  \\ \midrule
SSAT              & No       & 91.30   & 0.8318 & 21.84 & 0.2746 & 0.1810 & 6.81  \\
IP2P              & No       & 58.16  & 0.7870  & 21.69 & 0.2073 & 0.0995 & 6.88 \\ \midrule
ControlNet        & No       & 199.94 & 0.4381 & 21.31 & 0.6290  & 0.4395 & 6.94 \\
ControlNet+Text & GT       & 83.95 & 0.7774 & 21.92 & 0.3515 & 0.1608 & 4.17 \\ \midrule
Ours w/o H-AdaIN   & GT       & 44.83  & 0.8478 & 26.88 & 0.2233 & 0.0963 & / \\
Ours w/o H-AdaIN   & Detector & 44.97  & 0.8440  & 26.57 & 0.2285 & 0.0979 & / \\
Ours              & GT       & \textbf{\textcolor{red}{23.48}} & \textbf{\textcolor{red}{0.8556}} & \textbf{\textcolor{red}{28.24}} & \textbf{\textcolor{red}{0.1677}} & \textbf{\textcolor{red}{0.0737}} & \textbf{\textcolor{red}{2.58}} \\
Ours     & Detector & \textcolor{blue}{23.60}  & \textcolor{blue}{0.8532} & \textcolor{blue}{28.19} & \textcolor{blue}{0.1693} & \textcolor{blue}{0.0741} & \textcolor{blue}{2.72} \\ \bottomrule
\end{tabular}
\end{table*}
\section{Experiments}

\subsection{Implementation Details}
\label{Implement Details}
\textbf{Training details.} We choose RetouchingFFHQ~\cite{Ying_Liu_Li_Xu_Qian_Zhang_2023} as our training dataset.
We eliminate images containing closed eyes and infants and resize them to 512$\times$512. Finally, we used 19,757 images for training.
These images cover three retouching methods (enlarging eyes, face lifting, and smoothing skin) with different levels (ranging from 0 to 3) and include images with single or mixed retouching operations. 

The facial retouching detector 
is trained for 30 epochs on the entire training dataset, 
with the learning rate of $1\times10^{-5}$. 
For the FaceR, we follow the setting of vanilla ControlNet~\cite{zhang2023adding}, using the SD1.5. 
We train FaceR with a batch size of 20 and a learning rate of $1\times10^{-5}$ for 100 epochs.

\noindent\textbf{Evaluation setting.} We randomly select 1000 pairs of retouched and original images with different retouching methods and levels from the RetouchingFFHQ validation set for evaluation. We set up two sets of retouching labels: ground truth and the detection results. 

\subsection{Comparison with Other Methods}
\textbf{Comparing methods.} We select four methods for comparison, including SSAT~\cite{sun2022ssat}, InstructPix2Pix (IP2P)~\cite{brooks2023instructpix2pix}, ControlNet and ControlNet+Text. The first two are pre-existing methods without specific training on our task; the latter two are baseline methods, specifically trained on our task.

\begin{table}
\centering
\setlength{\tabcolsep}{8pt}
\caption{Ablation of H-AdaIN's hyperparameters. All experiments use retouching labels generated by the detector.
}

\label{tab:hadain}
\begin{tabular}{@{}ccccccc@{}}
\toprule
L   & $\gamma$       & FID$\downarrow$    & SSIM$\uparrow$ &PSNR$\uparrow$  & LPIPS$\downarrow$  & DISTS$\downarrow$  \\ \midrule
1  &0    & 44.06   & 0.8385 & 26.09   & 0.2312 & 0.0986 \\
30  &0   & 23.83   & 0.8523 & 28.15   & 0.1742 & 0.0774 \\
100  &0  & \textbf{\textcolor{red}{19.71}}   & \textbf{\textcolor{red}{0.8720}} & \textbf{\textcolor{red}{28.51}}   & \textbf{\textcolor{red}{0.1550}} & \textcolor{blue}{0.0745} \\ 
30  &0.5 & \textcolor{blue}{23.37}   & 0.8514 & 28.16   & 0.1723 & 0.0752 \\
30  &0.9 & 30.99   & 0.8501 & 27.76   & 0.1862 & 0.0851 \\ \midrule
30  &0.7 & 23.60   & \textcolor{blue}{0.8532} & \textcolor{blue}{28.19} & \textcolor{blue}{0.1693} & \textbf{\textcolor{red}{0.0741}} \\ \bottomrule
\end{tabular}
\end{table}

\begin{figure}
\centering
\includegraphics[width=0.47\textwidth]{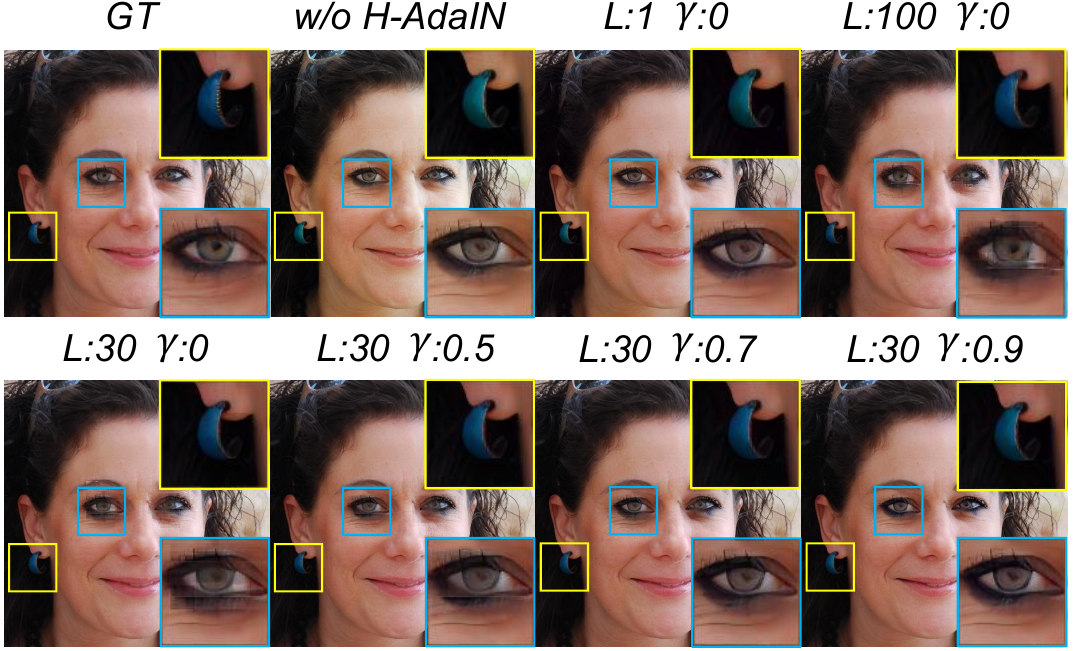}
\caption{The visualization results of applying different hyperparameters to H-AdaIN, "$L$" denotes hierarchical level, "$\gamma$" represents the overlapping ratio.} 
\label{fig:hadaincom}
\end{figure}

Originally designed for makeup transfer, SSAT requires the provision of both a makeup-applied image and a non-makeup image. In our experiments, we supply retouched photos alongside non-retouched photos as input, seeking to enable the model to transition the subject from a "retouched" state to a "non-retouched" state. IP2P is initially intended for instruction-based image editing; in our case, we introduce retouching degrees via textual prompts. ControlNet performs the blind reversion of retouched images without any prompt. ControlNet+Text introduces the retouching degrees through textual prompts. The approach to incorporating the retouching degrees is discussed in Sec.~\ref{Implement Details}. For methods that only support text conditions, we use text prompts instead of retouching labels. For example, "Revert the retouched face with EyeEnlarging 3, FaceLifting 1, Smoothing 0" corresponds to the retouching label "[3, 1, 0]".

\noindent \textbf{Evaluation metrics.} We select several commonly used evaluation metrics for image reversion tasks, including Peak Signal-to-Noise Ratio (PSNR) and Structural Similarity Index Measure (SSIM)~\cite{wang2004image}, as well as learning-based metrics, namely Learned Perceptual Image Patch Similarity (LPIPS)~\cite{zhang2018unreasonable}, Deep Image Structure and Texture Similarity (DISTS)~\cite{ding2020image} and Fréchet Inception Distance (FID)~\cite{heusel2017gans}. We also use HRNet~\cite{SunXLW19} for landmark prediction, and by calculating the average distance (AvgD) between the corresponding landmarks of the retouched image and the reverted image, the degree of reversion of the shape deformation can be quantified.

\noindent\textbf{Quantitative experiments.}
We present a quantitative comparison of various approaches on the RetouchingFFHQ test set. As shown in Tab.~\ref{tab:comp}, our method outperforms in all evaluation metrics. 
Specifically, Re-Face achieves an FID score of 23.48, which is 59.43$\%$ lower than IP2P 71.89$\%$ lower than ControlNet+Text, and 74.16$\%$ lower than SSAT. 
The AvgD of SSAT and IP2P is too similar to the retouched images, which can be considered incapable of reverting shape deformation. ControlNet+Text shows a noticeable improvement, but it still can't revert the shape very well.
Furthermore, Re-Face surpasses the compared methods in the remaining metrics, indicating its superiority. Including the GT label comparison in the experiment with Re-Face using a facial retouching detector aims to highlight the quality upper bound of Re-Face.

\noindent\textbf{Qualitative experiments.}
We conduct qualitative tests on the RetouchingFFHQ test set to demonstrate the effectiveness of our method, as shown in Fig. ~\ref{fig:compare_fig}. It is observed that Re-Face is capable of faithfully and naturally reverting images according to the retouching labels, as illustrated in the fifth column of Fig. ~\ref{fig:compare_fig}. In contrast, other approaches either exhibit no response to the instructions for retouching reversion or transform styles (as in the case of IP2P) or are capable of following the retouching label for reversion (such as ControlNet+Text) but result in unnatural coloration and details. Particularly, in the case of the smoothing operation, SSAT fails to handle this operation, whereas IP2P and ControlNet+Text, although able to identify and address smoothing, generate details that do not belong to the original image.

\subsection{Ablation Study on H-AdaIN}
We conduct comparative experiments on the hyperparameters involved in H-AdaIN, including the hierarchical level $L$ and overlapping ratio $\gamma$. The $L$ determines the maximum number of patches segmented from the original image, whereas the $\gamma$ governs the proportion of edges that overlap between adjacent patches. As a control, we also present the result employing AdaIN (i.e. $L$=1, $\gamma$=0).


When only AdaIN is used, the earring appears green as opposed to the ground truth of the blue, indicating that the color is not faithfully reverted. 
With $L$ set to 100, although the earring's color is faithfully reverted and the quantitative results surpass all other experiments, an erroneous enlargement of small eyes is discernible in Fig. ~\ref{fig:hadaincom}. This is because the H-AdaIN incorrectly applies pixel data from small patches that were part of the eye pre-reversion to non-eye regions post-reversion. Thus, a moderate $L$=30 is selected. 

Tab.~\ref{tab:hadain} reveals that when $L$ is set to 30 and $\gamma$ is less than 0.9, the differences in quantitative results are negligible. 
Fig. ~\ref{fig:hadaincom} offers additional insights: When $\gamma$ is too small, due to the lack of overlap in the processing of adjacent patches, conspicuous patch boundaries are apparent within the image. Conversely, when $\gamma$ is too large, H-AdaIN tends towards the behavior of AdaIN, thus experiencing similar issues with AdaIN. 
Ultimately, for $\gamma$, a moderate value of 0.7 is chosen.

\section{Conclusion}
We propose a framework for the reversion of retouched facial images. In addition, we propose a color correction module, Hierarchical Adaptive Instance Normalization (H-AdaIN), to address the issue of color shift. Extensive experiments demonstrate that our approach outperforms previous methods significantly in both qualitative and quantitative results. Our framework establishes a new paradigm for retouched image reversion, paving the way for further developments.

\bibliographystyle{IEEEbib}
\bibliography{1288}

\end{document}